%% file: emnlp2020.tex
\documentclass[11pt,a4paper]{article}
\usepackage[hyperref]{emnlp2020}
\usepackage{times}
\usepackage{latexsym}
\usepackage{graphicx}
\usepackage{subfigure}
\usepackage{booktabs} % for professional tables
\usepackage{xspace}
\usepackage{ctable}
\usepackage{amssymb}
\usepackage{amsmath}
\usepackage{color,soul}
\usepackage{amsthm}
\usepackage{multirow}
\usepackage{makecell}
\usepackage{enumitem}
\usepackage{mathtools}
\usepackage{calc}
\usepackage{hyperref}

\usepackage{microtype}

\aclfinalcopy % Uncomment this line for the final submission
\newcommand{\hvec}[1]{| {#1} \rangle}
\newcommand{\hcovec}[1]{\langle {#1} |}
\newcommand{\hdot}[2]{\langle {#1} | {#2} \rangle}
\newcommand{\vocab}{\mathcal{V}}
\newcommand{\corpus}{\mathcal{D}}

\newcommand{\hder}{\frac{\partial f_{ij}}{\partial\psi_{ij}}\xspace}
\newcommand{\hdersimple}{\frac{\partial f_{ij}}{\partial\hdot{i}{j}}\xspace}
\newcommand{\pmi}{\mathrm{PMI}\xspace}

\newcommand{\rank}[1]{\mathrm{rank}({#1})}

\DeclarePairedDelimiter\norm{\lVert}{\rVert}%

\title{Deconstructing word embedding algorithms}

\author{Kian Kenyon-Dean$^{\dagger}$\thanks{$\,\,\,$ Kian and Edward contributed equally. $\dagger$ This work was pursued while Kian was a member of Mila.} \\
  BMO AI Capabilities Team \\
  Bank of Montreal - Toronto, Ontario \\
  \texttt{kian.kenyon-dean@bmo.com} \\\And
  Edward Newell$^{*}$, Jackie Chi Kit Cheung \\
  Mila - Qu\'ebec AI Institute \\
  McGill University - Montr\'eal, Qu\'ebec \\
  \texttt{edward.newell@gmail.com}\\\texttt{jcheung@cs.mcgill.ca} \\}
  
\date{}

\begin{document}
\maketitle

\begin{abstract}
Word embeddings are reliable feature representations of words used to obtain high quality results for various NLP applications. Uncontextualized word embeddings are used in many NLP tasks today, especially in resource-limited settings where high memory capacity and GPUs are not available.
Given the historical success of word embeddings in NLP, we propose a retrospective on some of the most well-known word embedding algorithms. 
In this work, we deconstruct \textit{Word2vec}, \textit{GloVe}, and others, into a common form, unveiling some of the common conditions that seem to be required for making performant word embeddings. 
% Despite the differences that appear on the surface of these algorithms, we find that each one: (1) fits vector-covector dot products to approximate pointwise mutual information (PMI); and, (2) modulates the loss gradient to balance weak and strong signals. 
We believe that the theoretical findings in this paper can provide a basis for more informed development of future models.
\end{abstract}

\section{Introduction}
% uncontextualized embeddings are important in
% cognitive and psycho-linguistic research (and even in more general research
% of cognition, which dated back at least to the seminal Tversky work of
% 1977).

The advent of efficient uncontextualized word embedding algorithms (e.g., Word2vec \cite{mikolov2013distributed} and GloVe \cite{pennington2014glove}) marked a historical breakthrough in NLP. Countless researchers employed word embeddings in new models to improve results on a multitude of NLP problems. 
% Today, a similar shift seems to be happening with the advent of contextualized models; e.g., \citet{peters2018deep,devlin2018bert}.
In this work, we provide a retrospective analysis of these groundbreaking models of the past, which simultaneously offers theoretical insights for how future models can be developed and understood. 
We build on the theoretical work of \citet{levy2014neural}, proving that their findings on the relationship between pointwise mutual information (PMI) and word embeddings go beyond Word2vec and singular value decomposition.
%Analogous to a watchmaker who curiously scrutinizes the mechanical components comprising her watches' oscillators, our aim is to uncover what makes word embeddings ``tick''. 

% Word embedding algorithms train two sets of embeddings: the \textit{vectors} (``input'' vectors) and the \textit{covectors} (``output'', or, ``context'' vectors). 
% One can training these vectors and covectors with a variety of performant algorithms: the sampling-based shallow neural network of SGNS \cite{mikolov2013distributed}, GloVe's weighted least squares over global corpus statistics \cite{pennington2014glove}, and matrix factorization methods \cite{levy2014neural,levy2015improving,shazeer2016swivel}.  

In particular, we generalize several word embedding algorithms into a common form by proposing the \textit{low rank embedder} framework.  We deconstruct each algorithm into its constituent parts, and find that, despite their many different hyperparameters, the algorithms collectively intersect upon the following two key design features. First, vector-covector dot products are learned to approximate PMI statistics in the corpus. Second, modulation of the loss gradient, directly or indirectly, is necessary to balance weak and strong signals arising from the highly imbalanced distribution of corpus statistics.

These findings can provide an informed basis for future development of both new embedding algorithms and deep contextualized models.

\section{Fundamental concepts}
We begin by formally defining embeddings, their vectors and covectors (also known as ``input'' and ``output'' vectors \cite{rong2014word2vec, nalisnick2016improving}), and pointwise mutual information (PMI).

\paragraph{Embedding.}
In general topology, an embedding is understood as an injective structure preserving map, $f: X \rightarrow Y$, between two mathematical structures $X$ and $Y$.  A word embedding algorithm ($f$) learns an inner-product space ($Y$) to preserve a linguistic structure within a reference corpus of text, $\corpus$ ($X$), based on a vocabulary, $\mathcal{V}$.  The structure in $\corpus$ is analyzed in terms of the relationships between words induced by their co-appearances, according to a certain definition of context.
In such an analysis, each word figures dually: (1) as a focal element inducing a local context; and (2) as elements of the local contexts induced by focal elements.
To make these dual roles explicit, we distinguish two copies of the vocabulary: the \textit{focal}, or \textit{term}, words $\vocab_T$, and the \textit{context} words $\vocab_C$.

Word embedding consists of two maps:
\begin{align*}\begin{split}
\vocab_C &\longrightarrow \mathbb{R}^{1 \times d}
\qquad
\vocab_T \longrightarrow \mathbb{R}^{d \times 1}
\\
i &\longmapsto \hcovec{i} \quad\;\:\:\,\qquad j\longmapsto \hvec{j}.
\end{split}\end{align*}
We use Dirac notation to distinguish \textit{vectors} $\hvec{j}$, associated to focal words, from \textit{covectors} $\hcovec{i}$, associated to context words.
In matrix notation, $\hvec{j}$ corresponds to a column vector and $\hcovec{i}$ to a row vector.  Their inner product is $\hdot{i}{j}$.
% ; this inner product completely characterizes the learned vector space. 
We later demonstrate that many word embedding algorithms, intentionally or not, learn a vector space where the inner product between a focal word $j$ and context word $i$ aims to approximate their PMI in the reference corpus: $\hdot{i}{j} \approx \pmi(i,j)$.

\paragraph{Pointwise mutual information (PMI).}
PMI is a commonly used measure of association in computational linguistics, and has been shown to be consistent and reliable for many tasks \cite{terra2003frequency}. It measures the deviation of the cooccurrence probability between two words $i$ and $j$ from the product of their marginal probabilities:
\begin{equation} \label{eq:pmi}
\begin{split}
    \pmi(i,j) := \ln \frac{p_{ij}}{p_i p_j}
    = \ln \frac{N N_{ij}}{N_i N_j},
\end{split}
\end{equation}
where $p_{ij}$ is the probability of word $i$ and word $j$ cooccurring (for some notion of cooccurrence), and where $p_i$ and $p_j$ are marginal probabilities of words $i$ and $j$ occurring.  The empirical PMI can be found by  replacing probabilities with corpus statistics. Words are typically considered to cooccur if they are separated by no more than $w$ words; $N_{ij}$ is the number of counted cooccurrences between a \textit{context} $i$ and a \textit{term} $j$; $N_i$, $N_j$, and $N$ are computed by marginalizing over the $N_{ij}$ statistics.

\input{comparison_table.tex}

\section{Word embedding algorithms} \label{sec:low-rank}
We will now introduce the  \textit{low rank embedder} framework for deconstructing word embedding algorithms, inspired by the theory of generalized low rank models \cite{udell2016generalized}. We unify several word embedding algorithms by observing them all from the common vantage point of their \textit{global loss function}. Note that this framework is used for theoretical analysis, not necessarily implementation.

The global loss function for a \textit{low rank embedder} takes the following form:
\begin{align}
\mathcal{L} &= \sum_{\mathclap{
    (i,j) \in 
    \vocab_C\!\times\!\vocab_T
}} f_{ij}\Big(
   \;\psi(\hcovec{i}, \hvec{j}),
   \;\;\phi(i,j)\;
\Big),
\label{eq:lre-loss}
\end{align}
where $\psi(\hcovec{i}, \hvec{j})$ is a kernel function of learned model parameters, and $\phi(i,j)$ is some scalar function (such as a measure of association based on how often $i$ and $j$ appear in the corpus); we denote these with $\psi_{ij}$ and $\phi_{ij}$ for brevity. As well, $f_{ij}$ are loss functions that take $\psi_{ij}$ and $\phi_{ij}$ as inputs; all $f_{ij}$ satisfy the property:
\begin{align}
\frac{\partial f_{ij}}{\partial \psi_{ij}} = 0
\quad\text{at}\quad
\psi_{ij} = \phi_{ij}.
\label{eq:deriv-constraint}
\end{align}

The design variable $\phi_{ij}$ is some function of \textit{corpus statistics}, and its purpose is to  quantitatively measure some relationship between words $i$ and $j$.
The design variable $\psi_{ij}$ is a function of \textit{model parameters} that aims to approximate $\phi_{ij}$; i.e., an embedder's fundamental objective is to learn $\psi_{ij} \approx \phi_{ij}$, and thus to train embeddings that capture the statistical relationships measured by $\phi_{ij}$.  The simplest choice for the kernel function $\psi_{ij}$, is to take $\psi_{ij} = \hdot{i}{j}$. But the framework allows any function that is symmetric and positive definite, allowing the inclusion of bias parameters (e.g. in GloVe) and subword parameterization (e.g. in FastText). 
% Intuitively, a low rank embedder is trying to directly fit a kernel function of model parameters ($\psi_{ij}$) to some statistical relationship between words ($\phi_{ij}$) of our choosing. 
We later demostrate that skip-gram with negative sampling takes  $\phi_{ij} := \pmi(i,j) - \ln k$ and $\psi_{ij} := \hdot{i}{j}$, and then learns parameter values that approximate $\hdot{i}{j}\approx \pmi(i,j) - \ln k$.
%
%
% Though the specific choice of $\phi_{ij}$ varies slightly, existing low rank embedders generally base $\phi_{ij}$ on cooccurrence
% of words within a linear window $w$ words wide.
% But it is worth pointing out that $\phi_{ij}$ can in principle be any pairwise relationship encoded as a scalar function of corpus statistics, and that it need not be symmetric.
%

To understand the range of models encompassed, it is helpful to see how the framework relates (but is not limited) to matrix factorization.  
Consider $\phi_{ij}$ as providing the entries of a matrix: $\mathbf{M} := \left[\phi_{ij}\right]_{ij}$.
For models that take $\psi_{ij} = \hdot{i}{j}$, we can write $\hat{\mathbf{M}} = \mathbf{WV}$, where $\mathbf{W}$ is defined as having row $i$ equal to $\hcovec{i}$, and $\mathbf{V}$ as having column $j$ equal to $\hvec{j}$.
Then, the loss function can be rewritten as:
\begin{align*}
\mathcal{L} 
%&= \sum_{\mathclap{
%    (i,j) \in 
%    \vocab_C\!\times\!\vocab_T
%}} f_{ij}\Big(
%   \hat{\mathbf{M}}_{ij},
%   \mathbf{M}_{ij}
%\Big),
%\\
&= \sum_{\mathclap{
    (i,j) \in 
    \vocab_C\!\times\!\vocab_T
}} f_{ij}\Big(
   (\mathbf{WV})_{ij},\;
   \mathbf{M}_{ij}
\Big).
\end{align*}
This loss function can be interpreted as matrix reconstruction error, because the constraint in Eq.~\ref{eq:deriv-constraint} means that the gradient goes to zero as $\mathbf{WV} \approx \mathbf{M}$.

% It should be noted that the framework admits algorithms that one might not normally consider as matrix factorizations per se.  SGNS is an example of an algorithm that \textit{implicitly} factorizes a matrix without explicitly representing it in memory, and we discuss links between implicit and explicit matrix factorization later (\S\ref{sec:densesparse}).

Selecting a particular low rank embedder instance requires key design choices to be made:
we must chose the embedding dimension $d$, the form of the loss terms $f_{ij}$, the kernel function $\psi_{ij}$, and the association function $\phi_{ij}$.  The derivative of $f_{ij}$ with respect to $\psi_{ij}$, which we call the \textit{characteristic gradient}, helps compare models because it exhibits the action of the gradient yet is symmetric in the parameters. In the Appendix we show how this derivative relates to gradient descent.
% Thus, we address a specific embedder by the tuple $\left(d, \hder, \psi_{ij}, \phi_{ij}, \right)$.
%

In the following subsections, we present the derivations of $\hder$, $\psi_{ij}$, and $\phi_{ij}$ for SVD \cite{levy2014neural,levy2015improving}, SGNS \cite{mikolov2013distributed}, FastText \citep{joulin2017bag}, GloVe \cite{pennington2014glove}, and LDS \citep{arora2016latent}. The derivation for Swivel \cite{shazeer2016swivel} as a low rank embedder is trivial, as it is already posed as a matrix factorization of PMI statistics. We summarize the derivations in Table~\ref{tab:embedder-form-comparison}.

\subsection{SVD as a low rank embedder} \label{sec:svd}
Singular value decomposition (SVD) of the positive-PMI (PPMI) matrix is used by \citet{levy2014neural,levy2015improving} to produce word embeddings that perform more or less equivalently to SGNS and GloVe. Converting the PMI matrix into PPMI is a trivial preprocessing step; $\phi$ is augmented according to a factor $\alpha = 0$ such that $\phi_{ij} = 0 \,\, \forall \phi_{ij} \leq \alpha$. We now prove why SVD of the PMI matrix results in word embeddings with dot products $\hdot{i}{j} \approx \pmi(i, j)$, noting that this proof naturally holds for all augmentations of $\phi$ according to the $\alpha$ factor, including PPMI.

\paragraph{Proof.} Truncated SVD provides an optimal solution to problem $\min_{D}\| D - A \|_{F}$ for some integer $K$ less than the dimensionality of matrix $A$ such that $\rank{D} = K$ \cite{udell2016generalized}. The solution is the truncated SVD of $A$ where $D = \sum_{k=1}^{K} \sigma_k u_k v_k^\intercal$ with $\sigma$ being the $k^{th}$ singular value and $u_k$ and $v_k$ as the $k^{th}$ left and right singular vectors. 

Within our framework, the truncated SVD of the PMI matrix thus solves the following loss function (note $A_{ij}=\phi_{ij}=\pmi(i,j)$):
\begin{equation}
\mathcal{L} = - \sum_{\mathclap{
    (i,j) \in 
    \vocab_C\!\times\!\vocab_T
}} 
\big(\psi_{ij} - \pmi(i,j))^2,
\end{equation}
where $\psi_{ij} = u_i^\intercal \Sigma v_j$. Allowing the square matrix of singular values $\Sigma$ to be absorbed into the vectors (as in \citet{levy2015improving}), we have $\hcovec{i} = u_i$ and $\hvec{j} = \Sigma v_j$. Thus, taking the derivative $\hder$ (noting that $f_{ij}$ here is simply the squared difference between $\psi_{ij}$ and $\phi_{ij}$) and setting it equal to zero we observe:
\begin{equation}
    \hdot{i}{j} = \pmi(i,j).
\end{equation}

\subsection{SGNS as a low rank embedder} \label{sec:sgns}

\citet{mikolov2013distributed} proposed skip-gram with negative sampling with the following loss function:
\begin{equation*}
\mathcal{L} = - \sum_{\mathclap{(i,j) \in D_2}}
\Big\{
\ln \sigma \hdot{i}{j} + \sum_{\ell=1}^k \mathbb{E} \Big[ 
\ln (1 - \sigma \hdot{i'_\ell}{j})
\Big]
\Big\},
\end{equation*}
where $\sigma$ is the logistic sigmoid function, $D_2$ is a \textit{list} containing each cooccurrence of a context-word $i$ with a focal word $j$ in the corpus, and the expectation is taken by drawing $i'_\ell$ from the (smoothed) unigram distribution to generate $k$ ``negative samples'' for a given focal-word  \cite{mikolov2013distributed}. We will demonstrate that SGNS is a low rank embedder with $\hdot{i}{j} \approx \pmi - \ln k$.

\paragraph{Proof.} We can transform the loss function by counting the number of times each pair occurs in the corpus, $N_{ij}$, and the number of times each pair is drawn as a negative sample, $N_{ij}^-$, while indexing the sum over the set $\vocab_C \!\times\!\vocab_T$:
\begin{equation*} 
\mathcal{L} = - \sum_{\mathclap{
    (i,j) \in 
    \vocab_C\!\times\!\vocab_T
}} 
\Big\{
N_{ij}\ln \sigma \hdot{i}{j} +  
N_{ij}^- \ln( 1- \sigma \hdot{i}{j})
\Big\}.
\end{equation*}
%This can be viewed as a loss-function computed over the elements of the matrix product $\textbf{VW}$ with elements $(\mathbf{VW})_{ij} = \hdot{i}{j}$.  
%The target matrix being factorized can be seen by examining the partial derivative of the loss function with respect to this inner product, $\hder$:
%

The global loss is almost in the required form for a low rank embedder (Eq.~\ref{eq:lre-loss}), and the appropriate setting for the model approximation function is $\psi_{ij} = \hdot{i}{j}$.
Calculating the partial derivative with respect to the model approximation function $\psi_{ij}$, following algebraic manipulation (using the identity $a \equiv (a+b)\sigma(\ln \frac{a}{b})$), we arrive at the following definition of the characteristic gradient for SGNS as a low rank embedder, where $\hder = \frac{\partial\mathcal{L}}{\partial\hdot{i}{j}}$:
\begin{align}
% \hder &= \frac{\partial\mathcal{L}}{\partial\hdot{i}{j}}
\frac{\partial\mathcal{L}}{\partial\hdot{i}{j}}
   &= N_{ij}^- \sigma \hdot{i}{j} - N_{ij} (1 - \sigma \hdot{i}{j}) \nonumber \\ 
   &= (N_{ij} + N_{ij}^-) \bigg[ \sigma \big( \hdot{i}{j} \big)
- \sigma \big( \ln\frac{N_{ij}}{N_{ij}^-} \big) \bigg] \nonumber \\
   &= (N_{ij} + N_{ij}^-) \bigg[ \sigma \big( \psi_{ij} \big)
- \sigma \big( \phi_{ij} \big) \bigg].
\label{eq:sgns-M}
\end{align}
This provides that the association function for SGNS is $\phi_{ij} = \ln(N_{ij}/N_{ij}^-)$, since the derivative will be equal to zero at that point (Eq.~\ref{eq:deriv-constraint}). However, recall that negative samples are drawn according to the unigram distribution (or a smoothed variant \cite{levy2015improving}). This means that $N_{ij}^- = kN_iN_j/N$. Therefore, in agreement with \citet{levy2014neural}, we find that: 
\begin{align}
    \phi_{ij} = \ln \frac{N_{ij} N}{N_i N_j k} = \pmi(i,j) - \ln k.
\end{align}

\subsection{FastText as a low rank embedder}
% \todo{Flesh this out more.}
Proposed by \citet{joulin2017bag}, FastText's motivation is orthogonal to the present work. Its purpose is to provide subword-based representation of words to improve vocabulary coverage and generalizability of word embeddings. Nonetheless, it can also be understood as a low rank embedder .

\paragraph{Proof.} FastText uses a loss function that is identical to SGNS except that the vector for each word is taken as the sum of embeddings for all character $n$-grams appearing in the word, with $3 \leq n \leq 6$. Therefore, define $\hvec{j}$ by $\hvec{j} \equiv \sum_{g \in z(j)} \hvec{g}$, where $\hvec{g}$ is the vector for $n$-gram $g$, and $z(j)$ is the set of $n$-grams in word $j$. 
Covectors are accorded to words directly, so need not be redefined.  The loss function and the derivation of entries for Table~\ref{tab:embedder-form-comparison} is then formally identical to those for SGNS. This provides that $\psi_{ij} = \hdot{i}{j}$, and, $\phi_{ij} = \pmi(i,j) - \ln k$.

\subsection{GloVe as a low rank embedder} \label{sec:glove}
GloVe was proposed as an algorithm halfway between sampling methods and matrix factorization \cite{pennington2014glove}. 
% Its efficiency came from the fact that it only performs a partial factorization of the $\ln N_{ij}$ matrix, only using elements $N_{ij} > 0$. 
Ignoring samples where $N_{ij}=0$, GloVe uses the following loss function:
\begin{align}
\begin{split}
\mathcal{L} = 
   \sum_{ij} h(N_{ij}) \Big(\hdot{i}{j} + b_i + b_j - \ln N_{ij} \Big)^2 \\
% h(N_{ij}) = 
%   \min\left(1, \left(\frac{N_{ij}}{N_\mathrm{max}}\right)^{\alpha}\right),\qquad
\end{split}
\end{align}
where $b_i$ and $b_j$ are learned bias parameters, and  $h(N_{ij})$ is a weighting function sublinear in $N_{ij}$.

GloVe can be cast as a low rank embedder by using the model approximation function as a kernel with bias parameters, and setting the association measure to simply be the objective:
\begin{align*}
\psi_{ij} = \big[
   \,\hcovec{i}_1 \, \cdots \, \hcovec{i}_d \: b_i \; 1\,
\big]
&\cdot
\big[
   \,\hvec{j}_1 \cdots \hvec{j}_d \:\: 1 \:\: b_j\:
\big]^\intercal,
\\
\text{and}\quad
\phi_{ij}&=\ln N_{ij}.
\end{align*}

\begin{figure}[t]
    \centering
    \includegraphics[width=\columnwidth]{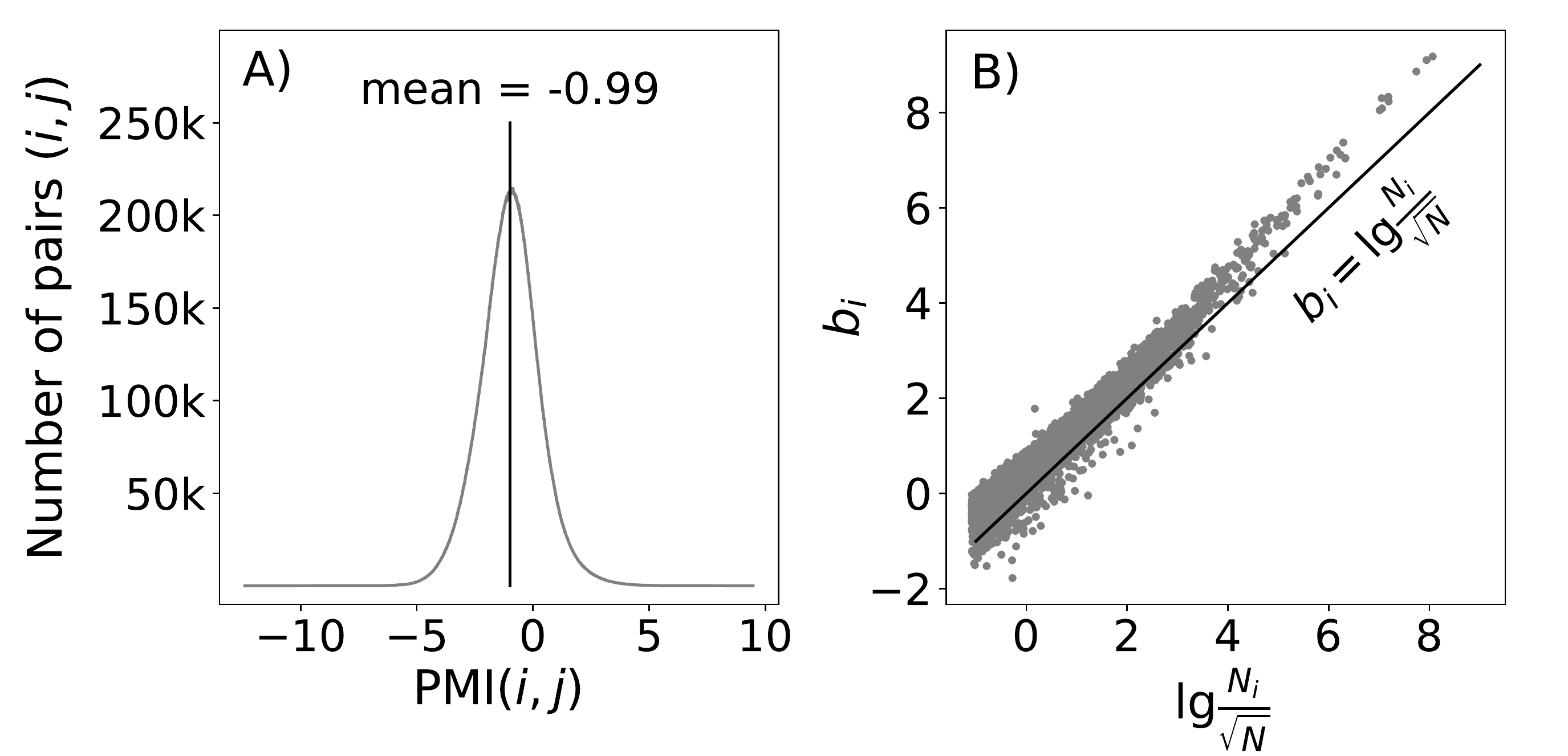}
    \caption{\textbf{A}) Histogram of $\pmi(i,j)$ values, for all pairs $(i,j)$ with $N_{ij}\!>\!0$. \textbf{B}) Scatter plot of GloVe's learned biases. Both from a Wikipedia 2018 corpus.}
    \label{fig:glove-is-pmi}
\end{figure}

\paragraph{Proof.} Observe an optimal solution to the loss function, when $\hder = 0$:
\begin{align*}
\hder &= 2h(N_{ij})\Big[\hdot{i}{j} + b_i + b_j - \ln N_{ij}\Big] = 0\\
&\implies
    \hdot{i}{j} + b_i + b_j = \ln N_{ij}.
\end{align*}
Multiplying the log operand by $1$:
\begin{align}
    \hdot{i}{j} + b_i &+ b_j = \ln \left(\frac{N_iN_j}{N}\frac{N}{N_iN_j}N_{ij}\right) \\
    &= \ln \frac{N_i}{\sqrt{N}} + \ln\frac{N_j}{\sqrt{N}} + \mathrm{PMI}(i,j) .
\label{eq:glove-is-pmi}
\end{align}
On the right side, we have two terms that depend respectively only on $i$ and $j$, which are candidates for the bias terms.  Based on this equation alone, we cannot draw any conclusions.  
However, empirically the bias terms are in fact very near $\frac{N_i}{\sqrt{N}}$ and $\frac{N_j}{\sqrt{N}}$, and
PMI is observed to be normally distributed, as can be seen in 
Fig.~\ref{fig:glove-is-pmi}.
This means that 
Eq.~\ref{eq:glove-is-pmi} provides $\hdot{i}{j} \approx \pmi(i,j)$.

Analyzing the optimum of GloVe's loss function yields important insights.  First, GloVe can be added to the list of low rank embedders that learn a bilinear parameterization of PMI.  Second, we can see why such a parameterization is advantageous.  Generally, it helps to standardize features of low rank models \cite{udell2016generalized}, and this is essentially what transforming cooccurrence counts into PMI achieves.  Thus, PMI can be viewed as a parameterization trick, providing an approximately normal target association to be modelled.

\subsection{LDS as a low rank embedder}
\citet{arora2016latent} introduced an embedding perspective based on generative modelling with random walks through a latent discourse space (LDS). LDS provided a theoretical basis for the performant \textit{SIF document embedding} algorithm, developed soon afterwards \cite{arora2017simple}. We now demonstrate that LDS is also a low-rank embedder.
% ote, however, that \citeauthor{arora2016latent}'s only experiments were on analogy tasks, which do not correlate well with performance on practical NLP problems \cite{linzen2016issues,faruqui2016problems,rogers2017too}, and despite including additional hyperparameters, their model showed no improvement over other algorithms. 

\paragraph{Proof.} The low rank learning objective for LDS follows directly from \textbf{Corollary 2.3}, in \citet{arora2016latent}:
\begin{equation*}
\begin{split}
    \pmi(i,j) &= \frac{\hdot{i}{j}}{d} + \gamma + O(\epsilon).
\end{split}
\end{equation*}
$\hder$ can be found by straightforward differentiation of LDS's loss function: 
\begin{equation*}
\mathcal{L} = \sum_{ij} h(N_{ij})\big[\ln N_{ij} - 
    \norm*{\hcovec{i}+\hvec{j}^\intercal}^2 -C \big]^2,
\end{equation*}
where $h(N_{ij})$ is as defined by GloVe.
The quadratic term is a valid kernel function because:
\begin{align*}
\hder = \norm*{\hcovec{i} + \hvec{j}^\intercal}^2 = 
\langle\tilde{i}|\tilde{j}\rangle,
\end{align*}
where
\begin{align*}
\tilde{\hcovec{i}} &= \Big[\sqrt{2}\hcovec{i}_1\:\,\cdots\;\sqrt{2}\hcovec{i}_d\;\;\hcovec{i}\hcovec{i}^\intercal\;\;\;\;1\;\;\;\;\;\Big]
,\\
\tilde{\hvec{j}} &= 
\Big[
\sqrt{2}\hvec{j}_1\;\cdots\;
\sqrt{2}\hvec{j}_d\;\;\;\,\;1\;\;\;\;\hvec{j}^\intercal\hvec{j}\;\Big]^\intercal.
\end{align*}

\section{Related work}
Our derivation of SGNS's solution is inspired by the work of \citet{levy2014neural}, who proved that \textit{skip-gram with negative sampling} (SGNS) \cite{mikolov2013distributed} was implicitly factorizing the $\pmi - \ln k$ matrix. However, they required additional assumptions for their derivation to hold.
\citet{li2015word} explored relations between SGNS and matrix factorization, but their derivation diverges from \citeauthor{levy2014neural}'s result and masks the connection between SGNS and other low rank embedders. Other works have also explored theoretical or empirical relationships between SGNS and GloVe \cite{shi2014linking,suzuki2015unified,levy2015improving,arora2016latent}.

\section{Discussion}
\label{sec:deconstructing}

% \todo{Add more meat to this section!}
We observe common features between each of the algorithms (Table~\ref{tab:embedder-form-comparison}). In each case, $\hder$ takes the form $(\mathrm{multiplier})\cdot(\mathrm{difference})$. 
The multiplier is always a ``tempered'' version of $N_{ij}$ (or $N_iN_j$); that is, it increases sublinearly with $N_{ij}$.

For each algorithm, $\phi_{ij}$ is equal to PMI or a scaled log of $N_{ij}$. Yet, the choice of $\psi_{ij}$ in combination with $\phi_{ij}$ provides that every model is optimized when $\hdot{i}{j}$ tends toward $\pmi(i,j)$ (with or without a constant shift or scaling). We demonstrated that the optimum for SGNS (and FastTest) is equivalent to the shifted PMI (\S\ref{sec:sgns}). For GloVe, we showed that incorporation of the bias terms captures the unigram counts needed for PMI (\S\ref{sec:glove}). A similar property is found in LDS with regards to the L2 norm in its learning objective \cite{arora2016latent}. Thus, these algorithms all converge on two key points: (1) an optimum in which model parameters are bilinearly related to PMI; and, (2) the weighting of $\hder$ by some tempered form of $N_{ij}$.

\section{Conclusion}
Our \textit{low rank embedder} framework has evoked the commonalities between many word embedding algorithms.  
We believe a robust understanding of these algorithms is a prerequisite for theoretically motivated development of deeper models. 
Indeed, we offer the following conjectures: deep embedding models would benefit by incorporating PMI statistics into their training objective; such models will also benefit from sub-linear scaling of frequent word pairs during training; and, lastly, such models would benefit by learning representations with a dual character, as all of the embedding algorithms we described do by learning vectors \textit{and} covectors.

% \todo{offer a conjecture for what these models could look like}

\section*{Acknowledgements}
This work is supported by the Fonds de recherche du Qu\'{e}bec -- Nature et technologies and by the Natural Sciences and Engineering Research Council of Canada. The last author is supported in part by the Canada CIFAR AI Chair program.

\bibliographystyle{acl_natbib}
\bibliography{emnlp2020}
% \newpage
\appendix

\section{Appendix}
\subsection{On the characteristic gradient}
% \todo{just show the update rules and justify why we can take a derivative w.r.t. the inner product}
The relationship between $\hder$ and the gradient descent actions taken during learning requires simply taking the next step in the chain rule during differentiation. For simplicity of exposition, we will assume, like SGNS and Swivel, that $\psi_{ij} = \hdot{i}{j}$, although the motivation of taking this derivative holds for any definition of $\psi_{ij}$, provided that it is a kernel function of the model parameters.

By examining the derivative $\hdersimple$ we observe the primary objective of the model (to approximate dot products), and how this objective symmetrically updates vectors and covectors during learning. 

Consider the generic update that occurs for a single $(i,j)$ pair with the pairwise loss function $f_{ij}$. The gradient descent rule for a single update to the vector for word $j$, using some learning rate $\eta$, is:
\begin{equation}
    \hvec{j} \leftarrow \hvec{j} - \eta \frac{\partial f_{ij}}{\hvec{j}},
\end{equation}
However, since $f_{ij}$ is a function of $\hdot{i}{j}$ and not of the vectors or covectors independently, we can use the chain rule to arrive at the following:
\begin{align}
    \hvec{j} & \leftarrow \hvec{j} - \eta \frac{\partial f_{ij}}{\partial\hdot{i}{j}}\frac{\partial \hdot{i}{j}}{\partial\hvec{j}}\\
    \hvec{j} & \leftarrow \hvec{j} - \eta \frac{\partial f_{ij}}{\partial{\hdot{i}{j}}} \hcovec{i}^{\intercal},
\end{align}
since $\frac{\partial \hdot{i}{j}}{\partial \hvec{j}} = \hcovec{i}$. Symmetrically, we also arrive at, for the updates to covectors:
\begin{equation}
    \hcovec{i} \leftarrow \hcovec{i} - \eta \frac{\partial f_{ij}}{\partial{\hdot{i}{j}}} \hvec{j}^{\intercal}.
\end{equation}
Therefore, taking $\hdersimple$ (more generally, $\hder$) to be the focal point of analysis in determining the objectives of the low rank embedders is well grounded. 
%Indeed, we also can observe that, at the objective when $\hder = 0$, gradient descent will clearly no longer update the model parameters, further suggesting that this point is indeed an optimium, and not just a saddle point.

\end{document}

%% file: comparison_table.tex
\setlength{\tabcolsep}{5pt}
\begin{table*}[t]
\centering
\begin{tabular}{c|l|c|c|c}
\textbf{Model} &
\multicolumn{1}{c|}{$\hder$}
& 
$\psi_{ij}$
&
$\phi_{ij}$
&
$\hdot{i}{j} \approx$
\rule{0pt}{0.8em}\rule[-0.8em]{0em}{0em}\\ \Xhline{3\arrayrulewidth}

%%%%%%%%%%%%% SVD row %%%%%%%%%%%%% 

SVD 

&

\rule{4.6em}{0em}
$2 \cdot \big[ \psi_{ij} - \phi_{ij} \big]$

& 

$\hdot{i}{j}$

&

$\pmi(i,j)$

&

$\pmi(i,j)$

\rule{0pt}{1.5em}\rule[-1.1em]{0em}{0em}\\ \hline

%%%%%%%%%%%%% SGNS row %%%%%%%%%%%%% 

SGNS 

&

\rule{0pt}{0em}
$(N_{ij} + N_{ij}^-) \cdot \big[ 
\sigma (\psi_{ij})
- \sigma(\phi_{ij})
\big]$

& 

$\hdot{i}{j}$

&

$\ln\frac{N_{ij}}{N_{ij}^-}$

&

$\pmi(i,j) - \ln k$

\rule{0pt}{1.5em}\rule[-1.1em]{0em}{0em}\\ \hline

%%%%%%%%%%%%% Glove row %%%%%%%%%%%%% 
GloVe

&

\rule{1.7em}{0em}
$2 h(N_{ij}) \cdot \big[\psi_{ij} - \phi_{ij} \big]\quad\;$

&

$\hdot{i}{j} + b_i + b_j$

& 

$\ln N_{ij}$

&

$\pmi(i,j)$

\rule{0pt}{1.5em}\rule[-1.0em]{0em}{0em}\\ \hline

%%%%%%%%%%%%% LDS row %%%%%%%%%%%%% 
LDS
& 
\rule{1.72em}{0em}
$4 h(N_{ij})
\cdot
\Big[
\psi_{ij}
-\phi_{ij} + C
\Big]$
&
$\norm{\hcovec{i}+\hvec{j}^\intercal}^2$
&
$\ln N_{ij}$
&
$d \pmi(i,j) - d\gamma$
\rule{0pt}{1.5em}\rule[-1.0em]{0em}{0em}\\ \hline

%%%%%%%%%%%%% Swivel row %%%%%%%%%%%%% 

\multirow{2}{*}{Swivel\rule{0em}{1.8em}}
& 
\rule{2.52em}{0em}
$\sqrt{N_{ij}} \cdot \Big[ \psi_{ij} - \phi_{ij} \Big]$
&
\multirow{2}{*}{
$\hdot{i}{j}$
\rule{0em}{1.8em}}
&
$\pmi(i,j)$
&
$\pmi(i,j)$

\rule{0pt}{1.5em}\rule[-1.0em]{0em}{0em}\\ 
& 
\rule{4.48em}{0em}
$1\cdot \sigma \Big(\psi_{ij} - \phi_{ij} \Big)$
&
&
$\pmi^*(i,j)$
&
$\pmi^*(i,j)$
% \rule{0pt}{0em}\rule[-.8em]{0em}{0em}\\[2mm] \Xhline{3\arrayrulewidth}

%%%%%%%%%%%%% Hilby row %%%%%%%%%%%%% 

%\pbox{2cm}{\centering Hilbert-MLE}
% \hilby

% & 
% \rule{1.69em}{0em}
% $\big(p_i p_j\big)^\frac{1}{\tau} \cdot \Big[
% e^{\psi_{ij}}
% - e^{\phi_{ij}}
% \Big]$
% &
% $\hdot{i}{j}$
% &
% $\pmi(i,j)$
% &
% $\pmi(i,j)$

\rule{0pt}{1.5em}\rule[-0em]{0em}{0em}\\ 

    \end{tabular}
    \caption{Comparison of low rank embedders.  Final column shows the value of $\hdot{i}{j}$ at $\hder=0$.
    GloVe and LDS set $f_{ij}=0$ when $N_{ij}=0$; $h(N_{ij})$ is a weighting function sublinear in $N_{ij}$.  Swivel takes one form when $N_{ij}>0$ (first row) and another when $N_{ij}=0$ (second row).
    $N_{ij}^-$ is the number of negative samples;
    in SGNS, $N_{ij}^- \propto N_iN_j$, and both $N_{ij}$ and $N_{ij}^-$ are tempered by undersampling and unigram smoothing.}%: SGNS \cite{mikolov2013distributed}, GloVe \cite{pennington2014glove}, LDS \cite{arora2016latent}, Swivel \cite{shazeer2016swivel}.}
    \label{tab:embedder-form-comparison}
\end{table*}

%% file: emnlp2020.bbl
\begin{thebibliography}{15}
\expandafter\ifx\csname natexlab\endcsname\relax\def\natexlab#1{#1}\fi

\bibitem[{Arora et~al.(2016)Arora, Li, Liang, Ma, and
  Risteski}]{arora2016latent}
Sanjeev Arora, Yuanzhi Li, Yingyu Liang, Tengyu Ma, and Andrej Risteski. 2016.
\newblock A latent variable model approach to pmi-based word embeddings.
\newblock \emph{Transactions of the Association for Computational Linguistics},
  4:385--399.

\bibitem[{Arora et~al.(2017)Arora, Liang, and Ma}]{arora2017simple}
Sanjeev Arora, Yingyu Liang, and Tengyu Ma. 2017.
\newblock A simple but tough-to-beat baseline for sentence embeddings.
\newblock \emph{International Conference on Learning Representations}.

\bibitem[{Joulin et~al.(2017)Joulin, Grave, and Mikolov}]{joulin2017bag}
Armand Joulin, Edouard Grave, and Piotr Bojanowski~Tomas Mikolov. 2017.
\newblock Bag of tricks for efficient text classification.
\newblock \emph{European Association for Computational Linguistics 2017}, page
  427.

\bibitem[{Levy and Goldberg(2014)}]{levy2014neural}
Omer Levy and Yoav Goldberg. 2014.
\newblock Neural word embedding as implicit matrix factorization.
\newblock In \emph{Advances in Neural Information Processing Systems}, pages
  2177--2185.

\bibitem[{Levy et~al.(2015)Levy, Goldberg, and Dagan}]{levy2015improving}
Omer Levy, Yoav Goldberg, and Ido Dagan. 2015.
\newblock Improving distributional similarity with lessons learned from word
  embeddings.
\newblock \emph{Transactions of the Association for Computational Linguistics},
  3:211--225.

\bibitem[{Li et~al.(2015)Li, Xu, Tian, Jiang, Zhong, and Chen}]{li2015word}
Yitan Li, Linli Xu, Fei Tian, Liang Jiang, Xiaowei Zhong, and Enhong Chen.
  2015.
\newblock Word embedding revisited: a new representation learning and explicit
  matrix factorization perspective.
\newblock In \emph{Proceedings of the 24th International Conference on
  Artificial Intelligence}, pages 3650--3656. AAAI Press.

\bibitem[{Mikolov et~al.(2013)Mikolov, Sutskever, Chen, Corrado, and
  Dean}]{mikolov2013distributed}
Tomas Mikolov, Ilya Sutskever, Kai Chen, Greg~S Corrado, and Jeff Dean. 2013.
\newblock Distributed representations of words and phrases and their
  compositionality.
\newblock In \emph{Advances in neural information processing systems}, pages
  3111--3119.

\bibitem[{Nalisnick et~al.(2016)Nalisnick, Mitra, Craswell, and
  Caruana}]{nalisnick2016improving}
Eric Nalisnick, Bhaskar Mitra, Nick Craswell, and Rich Caruana. 2016.
\newblock Improving document ranking with dual word embeddings.
\newblock In \emph{Proceedings of the 25th International Conference Companion
  on World Wide Web}, pages 83--84. International World Wide Web Conferences
  Steering Committee.

\bibitem[{Pennington et~al.(2014)Pennington, Socher, and
  Manning}]{pennington2014glove}
Jeffrey Pennington, Richard Socher, and Christopher Manning. 2014.
\newblock Glove: Global vectors for word representation.
\newblock In \emph{Proceedings of the 2014 conference on Empirical Methods in
  Natural Language Processing}, pages 1532--1543.

\bibitem[{Rong(2014)}]{rong2014word2vec}
Xin Rong. 2014.
\newblock Word2vec parameter learning explained.
\newblock \emph{arXiv preprint arXiv:1411.2738}.

\bibitem[{Shazeer et~al.(2016)Shazeer, Doherty, Evans, and
  Waterson}]{shazeer2016swivel}
Noam Shazeer, Ryan Doherty, Colin Evans, and Chris Waterson. 2016.
\newblock Swivel: Improving embeddings by noticing what's missing.
\newblock \emph{arXiv preprint arXiv:1602.02215}.

\bibitem[{Shi and Liu(2014)}]{shi2014linking}
Tianze Shi and Zhiyuan Liu. 2014.
\newblock Linking glove with word2vec.
\newblock \emph{arXiv preprint arXiv:1411.5595}.

\bibitem[{Suzuki and Nagata(2015)}]{suzuki2015unified}
Jun Suzuki and Masaaki Nagata. 2015.
\newblock A unified learning framework of skip-grams and global vectors.
\newblock In \emph{Proceedings of the 53rd Annual Meeting of the ACL and the
  7th IJCNLP (Volume 2: Short Papers)}, volume~2, pages 186--191.

\bibitem[{Terra and Clarke(2003)}]{terra2003frequency}
Egidio Terra and Charles~LA Clarke. 2003.
\newblock Frequency estimates for statistical word similarity measures.
\newblock In \emph{Proceedings of the 2003 Conference of NAACL-HLT - Volume 1},
  pages 165--172. Association for Computational Linguistics.

\bibitem[{Udell et~al.(2016)Udell, Horn, Zadeh, Boyd
  et~al.}]{udell2016generalized}
Madeleine Udell, Corinne Horn, Reza Zadeh, Stephen Boyd, et~al. 2016.
\newblock Generalized low rank models.
\newblock \emph{Foundations and Trends in Machine Learning}, 9(1):1--118.

\end{thebibliography}
